\PassOptionsToPackage{table}{xcolor}
\documentclass[10pt,twocolumn,letterpaper]{article}

\usepackage[pagenumbers]{cvpr} %

\usepackage{colortbl}
\definecolor{bestcell}{RGB}{255,255,255}  %

\definecolor{cvprblue}{rgb}{0.21,0.49,0.74}
\usepackage[pagebackref,breaklinks,colorlinks,allcolors=cvprblue]{hyperref}

\def\paperID{} %
\def\confName{CVPR}
\def\confYear{2026}

\title{MatSpray: Fusing 2D Material World Knowledge on 3D Geometry}

\author{Philipp Langsteiner\\
{\tt\small philipp.langsteiner@uni-tuebingen.de}
\and
Jan-Niklas Dihlmann\\
{\tt\small jan-niklas.dihlmann@uni-tuebingen.de}
\and
Hendrik Lensch\\
{\tt\small hendrik.lensch@uni-tuebingen.de}
}

\begin{document}
\twocolumn[{%
\maketitle
\vspace{-2em}
\begin{center}
    \includegraphics[width=\textwidth]{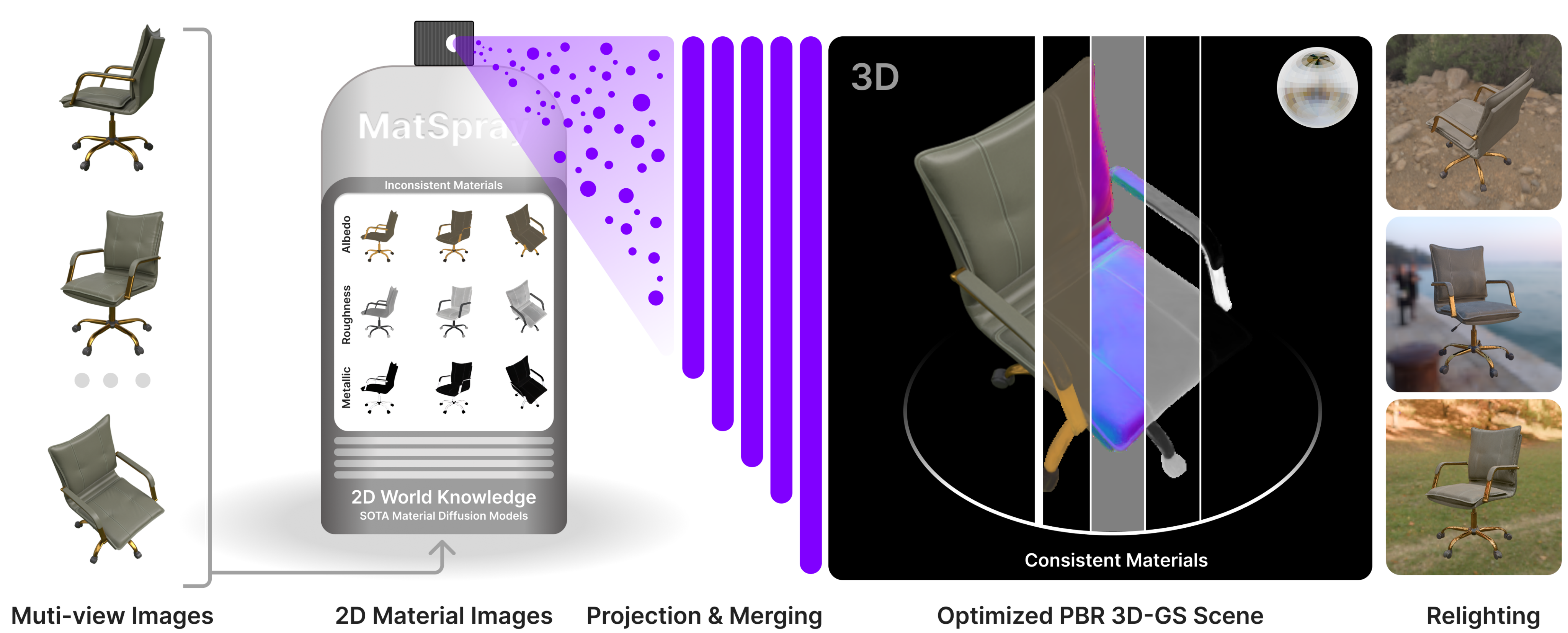}
    \captionof{figure}{\textbf{MatSpray Overview} we utilize 2D material world knowlegde from 2D diffusion models to reconstruct 3D relightable objects. Given multi-view images of a target object, we first generate per-view PBR material predictions (base color, roughness, metallic) using any 2D diffusion-based material model. These 2D estimates are then integrated into a 3D Gaussian Splatting reconstruction via Gaussian ray tracing. Finally, a neural refinement stage applies a softmax-based restriction to enforce multi-view consistency and enhance the physical accuracy of the materials. The resulting 3D assets feature high-quality, fully relightable PBR materials under novel illumination. Project page: \protect\url{https://matspray.jdihlmann.com/}}
    \label{fig:teaser}
\end{center}
\vspace{1em}
}]
\begin{abstract}
Manual modeling of material parameters and 3D geometry is a time consuming yet essential task in the gaming and film industries. While recent advances in 3D reconstruction have enabled accurate approximations of scene geometry and appearance, these methods often fall short in relighting scenarios due to the lack of precise, spatially varying material parameters. At the same time, diffusion models operating on 2D images have shown strong performance in predicting physically based rendering (PBR) properties such as albedo, roughness, and metallicity. However, transferring these 2D material maps onto reconstructed 3D geometry remains a significant challenge. We propose a framework for fusing 2D material data into 3D geometry using a combination of novel learning-based and  projection-based approaches. We begin by reconstructing scene geometry via Gaussian Splatting. From the input images, a diffusion model generates 2D maps for albedo, roughness, and metallic parameters. Any existing diffusion model that can convert images or videos to PBR materials can be applied.  The predictions are further integrated into the 3D representation either by optimizing an image-based loss or by directly projecting the material parameters onto the Gaussians using Gaussian ray tracing. To enhance fine-scale accuracy and multi-view consistency, we further introduce a light-weight neural refinement step (Neural Merger), which takes ray-traced material features as input and produces detailed adjustments. Our results demonstrate that the proposed methods outperform existing techniques in both quantitative metrics and perceived visual realism. This enables more accurate, relightable, and photorealistic renderings from reconstructed scenes, significantly improving the realism and efficiency of asset creation workflows in content production pipelines. Project page: \protect\url{https://matspray.jdihlmann.com/}
\end{abstract}
    
\section{Introduction}
\label{sec:intro}

Editing and relighting real scenes captured with casual cameras is central to many vision and graphics applications. While modern neural 3D reconstruction methods can produce impressive geometry and appearance from images, they often entangle illumination with appearance, yielding textures or coefficients that are not physically meaningful for relighting. Classical inverse rendering requires strong assumptions about lighting and exposure and remains fragile when materials vary spatially. In parallel, recent 2D material predictors learn rich priors from large-scale data and can produce plausible material maps from images, yet they operate in 2D and are not directly consistent across views or attached to a 3D representation.

We introduce a method to transfer 2D material predictions onto a 3D Gaussian representation to obtain relightable assets with spatially varying base color, roughness, and metallic parameters. The approach projects 2D material maps to 3D via efficient ray-traced assignment, refines materials with a small MLP to reduce multi-view inconsistencies, and supervises rendered material maps directly with the 2D predictions to preserve plausible priors while discouraging baked-in lighting. This combination yields cleaner albedo, more accurate roughness, and informed metallic estimates, enabling higher-quality relighting compared to pipelines that learn only appearance. 
Our contributions are:
\begin{itemize}
  \item \textbf{World Material Fusion.} A plug-and-play pipeline that, to our knowledge, is the first to fuse swappable diffusion-based 2D PBR priors (“world material knowledge”) with 3D Gaussian material optimization via Gaussian ray tracing and PBR consistent supervision to obtain relightable assets.
  \item \textbf{Neural Merger.} A softmax neural merger that aggregates per-Gaussian, multi-view material estimates, suppresses baked-in lighting, and enforces cross-view consistency while stabilizing joint environment map optimization.
  \item \textbf{Faster Reconstruction.} A simple projection and optimization scheme that reconstructs high-quality relightable 3D materials with \textbf{3.5$\times$} less per-scene optimization time than IRGS \cite{gao:24}. 
\end{itemize}

\section{Related Work}
\label{sec:related_work}

\paragraph{Materials}
Spatially varying BRDFs (svBRDFs) have long been studied, with early high-resolution texel-based representations enabling point-wise material parameterization \cite{Sato:97, Lensch:01}. In this work, we adopt a Cook–Torrance variant, which is based on the widely used Disney principled BRDF and real-time formulations in major engines \cite{burley:12, Karis:2013, Lagarde:14}. Building on this foundation, recent research has explored richer material parameterizations and extended the expressiveness of svBRDF models \cite{guo2025epbr}.

\paragraph{Diffusion}
Diffusion models enable high-fidelity image synthesis and conditioning, with efficient latent-space formulations and extensions to video generation \cite{ho:20, rombach:22, ho:22-video, blattmann:23}. For material estimation from 2D images, large-scale diffusion priors have been used to infer PBR maps \cite{zeng:24, ke:25, zheng2025dnf, liang:25, engelhardt2025-svim3d}. Of particular relevance is DiffusionRenderer by Huang et al. \cite{huang:25}, whose results on high-quality material maps motivated this work. Diffusion approaches have been further explored for related tasks such as HDR prediction, texture estimation, and relighting \cite{barua2025physhdrlightingmeetsmaterials, ying2025chordchainrenderingdecomposition, litman2025lightswitch}.

\paragraph{Gaussian Ray Tracing}
 We exploit Gaussian Ray Tracing as a mechanism for transferring 2D material data to 3D.
Although the field remains in its early stages, emerging works have explored stochastic and explicit Gaussian ray-tracing methods~\cite{sun2025stochastic, moenne-loccoz:24, mai2024ever} and related neural optimization schemes \cite{govindarajan2025radiant}. Our formulation builds upon insights from Mai et al.~\cite{mai2024ever} and Moenne-Loccoz et al.~\cite{moenne-loccoz:24}, extending them toward material-aware 3D reconstruction.

\begin{figure*}[t]
    \centering
    \includegraphics[width=\textwidth]{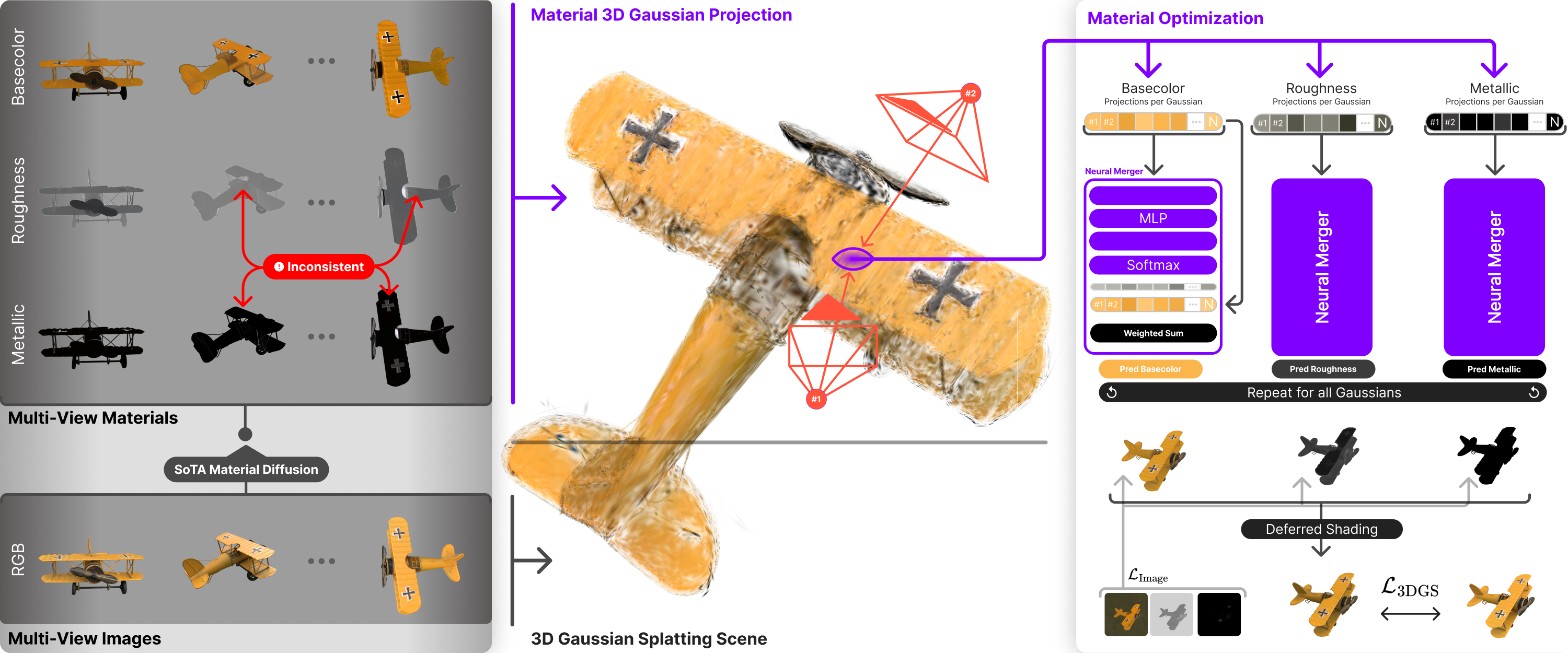}
    \caption{\textbf{Pipeline}. From multi-view images, a diffusion predictor yields per-view material maps. We reconstruct the object's geometry using 3D Gaussian Splatting. Then we project 2D materials to 3D via ray tracing, and refine per Gaussian materials with our Neural Merger that has a softmax output layer, choosing between the projected values. We then supervise the produced material maps using the predicted 2D material maps. Additionally, using deferred shading we supervise by a PBR-based photometric rendering loss with the multi-view ground truth images of the object.}
    \label{fig:method_overview_cvpr}
\end{figure*}

\paragraph{Novel View Synthesis and Scene Reconstruction}
Neural representations for novel view synthesis and scene reconstruction have rapidly advanced 3D modeling, with NeRF and Gaussian Splatting providing strong foundations for radiance-based scene representations \cite{mildenhall:21, kerbl:23}. Material modeling atop radiance fields has been explored through inverse rendering and relighting \cite{boss2021nerd, boss2021neural, zhang2021nerfactor, rudnev2022nerf}, where coupling with signed distance fields (SDFs) improves physical plausibility \cite{wu2025pbr, mao2025neus}. Gaussian-based inverse rendering approaches such as R3DGS \cite{gao:24} reconstruct materials via per-scene optimization, while IRGS \cite{gu:25} extends this with 2D Gaussians and deferred shading for improved appearance modeling. Complementary efforts leverage diffusion models for 3D reconstruction and sparse-view recovery \cite{mithun2025diffusion, liu2025dtnerfdiffusiontransformerbasedoptimization}, and hybrid cues or mesh integration further enhance geometric fidelity \cite{wang2025hyrf, xu2025depthsplat, liao2025rosgsrelightableoutdoorscenes, liang:24}. Beyond these, extensions of 2D and 3D Gaussians have enabled spatially varying materials, advanced relighting, and richer reflectance modeling \cite{sun:25, kouros2025rgs, yang2025gogshighfidelitygeometryrelighting, chen2024gi, li2025recap, hu2025real, kaleta2025lumigauss, zhao2024illuminerf, dihlmann:24, liu2024bigs, wu20253d, liu2025occlugaussian, lai2025glossygs}.

In contrast to R3DGS and IRGS, which optimize material and geometry parameters per scene, our approach employs Gaussian Ray Tracing to lift 2D material estimates by diffusion models into 3D representations, exploiting the world-knowledge priors learned by the diffusion models and the broader understanding of material behavior they provide. This enables faster, more consistent reconstruction and material reasoning across scenes.

\section{Method}

Our method recovers consistent 3D PBR materials from multiple views by combining 2D diffusion predictions with a 3D Gaussian representation. We first obtain material maps (base color, roughness, metallic) per view from any diffusion material predictor, making our approach compatible with a wide range of existing and future diffusion models. 
Scene geometry is reconstructed via a relightable Gaussian Splatting pipeline (R3DGS) \cite{kerbl:23, gao:24}, which provides both geometry and normals. The 2D material estimates are then transferred to 3D and jointly refined for multi-view consistency by a newly introduced Neural Merger step. 
The materials maps and normals are further refined based on a rendering loss, evaluated with deferred rendering. The illumination is modeled by an optimizable environment map during refinement. 

Specifically, we (1) lift 2D materials to 3D via Gaussian ray tracing, (2) refine per-Gaussian material parameters with the Neural Merger across views, and (3) supervise rendered material maps to preserve plausible 2D priors while suppressing baked-in lighting.

\subsection{Diffusion Material Prediction}
We leverage pretrained diffusion priors to predict physically meaningful per-view material maps, enabling accurate reconstruction of 3D PBR materials. Each material channel, base color (three channels), roughness (single channel), and metallic (single channel), is inferred explicitly. In practice, we evaluate multiple prebuilt diffusion-based predictors and select the one that provides the best fidelity and consistency balance for our data. In our experiments, this is DiffusionRenderer \cite{liang:25}. We also tested Marigold \cite{ke:25} and RGB-to-X \cite{zeng:24}. We choose DiffusionRenderer because it achieves about thirty percent higher PSNR than the other methods.

DiffusionRenderer predicts material maps from short frame batches and uses a limited temporal context to improve consistency and material understanding within each batch. While this improves local consistency, the predicted materials in a batch can still differ across views. The model cannot recover the complete environment illumination from a small input batch, which may lead to small shifts in color, roughness, or metallic appearance within a batch. In addition, results from separate batches may not align, as these images may introduce new information that was not visible in previous batches. These variations make a direct projection of the predicted maps unreliable and often result in blurry and washed out material maps. Our method resolves this by refining and merging the estimates into one consistent 3D representation.

\subsection{2D-to-3D Material Lifting}

For each view $v_i$, we collect the material attributes (base color, roughness, metallic) corresponding to every Gaussian $g$ from the pixels within its projected footprint $fp_p$ using Gaussian ray tracing, following the formulation of Mai et al.~\cite{mai:24}. Their approach determines each Gaussian or ellipsoid's contribution to a ray based on density. Because the opacity $\alpha$ used in Gaussian Splatting~\cite{kerbl:23} does not directly correspond to a physical density, we adopt the formulation by Moenne-Loccoz et al.~\cite{moenne-loccoz:24}, which allows the direct use of Gaussian Splatting opacity $\alpha$ in ray tracing.

For a Gaussian with mean $\boldsymbol{\mu}$ and covariance $\boldsymbol{\Sigma}$, the point of maximum response $\mathbf{x}_{\max}$ along a ray with origin $\mathbf{o}$ and direction $\mathbf{d}$ is
\begin{equation}
\tau_{\max} = \frac{(\boldsymbol{\mu} - \mathbf{o})^{\top} \boldsymbol{\Sigma}^{-1} \mathbf{d}}{\mathbf{d}^{\top} \boldsymbol{\Sigma}^{-1} \mathbf{d}}, \quad
\mathbf{x}_{\max} = \mathbf{o} + \tau_{\max} \mathbf{d}.
\end{equation}
The corresponding opacity $\alpha_{max}$, given a base opacity $\alpha$ and a falloff parameter $\lambda > 0$, is
\begin{equation}
\alpha_{\max} = \alpha \cdot \exp\left(-\frac{1}{2} \lambda (\mathbf{x}_{\max} - \boldsymbol{\mu})^{\top} \boldsymbol{\Sigma}^{-1} (\mathbf{x}_{\max} - \boldsymbol{\mu})\right).
\end{equation}

Material values per pixel $\mathbf{m}_{p}$ are then assigned to the Gaussians intersected by the ray and aggregated across all pixels in each Gaussian's footprint $\mathrm{fp}_{g,v_i}$ per view. To reduce color distortion from outliers and overlapping footprints, we compute a median of all assigned material parameters $\mathbf{m}_{g,v_i}$ per Gaussian:
\begin{equation}
\mathbf{m}_{g,v_i} = \mathrm{median}_{p \in \mathrm{fp}_{g,v_i}}(\mathbf{m}_{p}).
\end{equation}
After computing the median for each view, the resulting values are assigned to their corresponding Gaussians. Gaussians not intersected in any view are removed. Grid-based supersampling per pixel is used to ensure stable Gaussian hits.

For each Gaussian $g$, we now obtain arrays of material estimates across all views:
\begin{align}
    \text{basecolor}_g &= \{ b_{g,1},\, b_{g,2},\, \dots,\, b_{g,n} \}, \\
    \text{metallic}_g  &= \{ m_{g,1},\, m_{g,2},\, \dots,\, m_{g,n} \}, \\
    \text{roughness}_g &= \{ r_{g,1},\, r_{g,2},\, \dots,\, r_{g,n} \}.
\end{align}
These arrays contain the inconsistent material values per view produced by DiffusionRenderer. These inconsistencies motivate the subsequent Neural Merger step, which refines the estimates into a coherent 3D representation.

\subsection{Neural Merger}

To reduce inconsistencies across views, we introduce the \textit{Neural Merger}, which predicts weights per view for the material parameters collected during the projection step for each Gaussian. It fuses the predictions into a single, consistent estimate. The key idea is to interpolate between the predicted values rather than allowing the network to freely generate new colors or material values. This ensures that the merged results remain consistent with the world knowledge captured by the diffusion priors while enforcing coherence across views.

For each Gaussian $g$, the Neural Merger takes as input the projected material values $\mathbf{m}_{g,v}$ for all views $v \in \{1, \dots, V\}$, along with the Gaussian position $\mathbf{p}_g$, encoded using a positional encoding. The input is processed by a lightweight MLP $f_\theta$ to produce unnormalized weights $h_{g,v}$ for each view:
\begin{equation}
[h_{g,1}, h_{g,2}, \dots, h_{g,V}] = f_\theta \Big( \mathbf{p}_g, \mathbf{m}_{g,1}, \dots, \mathbf{m}_{g,V} \Big).
\end{equation}
A softmax function then converts these outputs into normalized weights:
\begin{equation}
w_{g,v} = \frac{\exp(h_{g,v})}{\sum_{v'=1}^{V} \exp(h_{g,v'})}, \quad \sum_{v=1}^{V} w_{g,v} = 1.
\end{equation}
The merged material $m_g$ for the Gaussian is computed as the weighted sum of the per-view predictions:
\begin{equation}
\mathbf{m}_g = \sum_{v=1}^{V} w_{g,v} \, \mathbf{m}_{g,v}.
\end{equation}
The Neural Merger is optimized during the refinement explained in the next section. 
Using the softmax weighting is crucial. Without it, the merger can converge faster than the environment map optimization, producing unrealistic material values that match the ground truth only superficially. By interpolating among the predicted values, the Neural Merger produces physically plausible material estimates while allowing the environment map to converge reliably. In our framework, we use a separate Neural Merger for each material channel, enabling improved disentanglement of the materials.

\begin{figure*}[t]
    \centering
    \includegraphics[width=\textwidth]{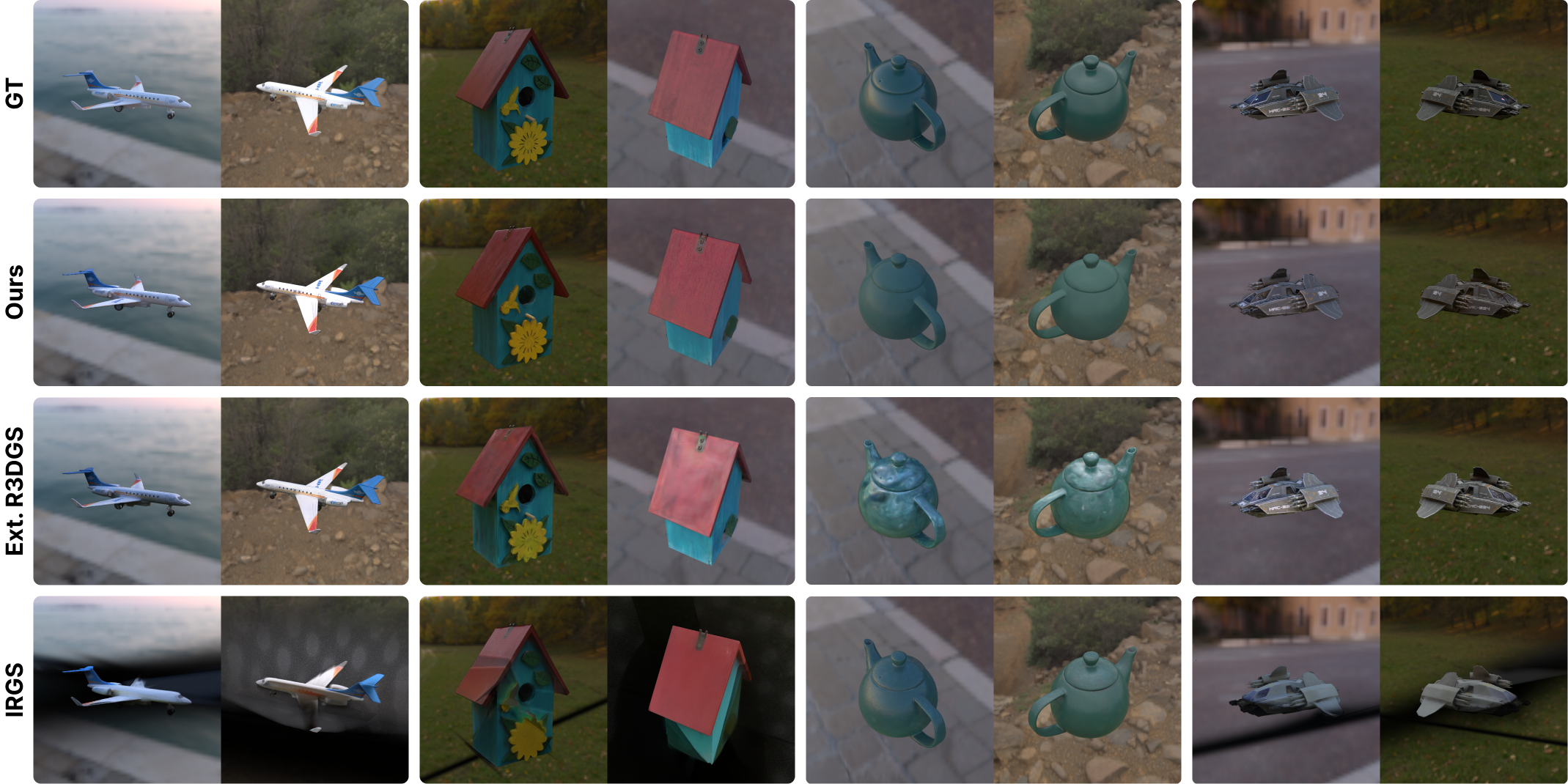}
    \caption{\textbf{Relighting Comparison} between our method, an extended version of R3DGS \cite{gao:24} %
    and IRGS \cite{gu:25}. The objects are all relit under the same environment maps. In IRGS, reconstructed scene geometry might partially occlude the environment map.}
    \label{fig:relighting}
\end{figure*}

\subsection{Refinement, Supervision and Loss Functions}

The Neural Merger produces material values per Gaussian, which are then rasterized into material maps. 
These maps are iteratively refined using two complementary supervised losses.

First, we supervise the rendered material maps against the diffusion model’s 2D material predictions using an $\mathcal{L}_1$ loss. This loss is applied exclusively to the material parameters, thereby optimizing only the Neural Merger. Given the rendered material maps $\mathbf{M}_{\text{render}}$ and the diffusion-predicted material maps $\mathbf{M}_{\text{2D}}$, the material supervision loss $\mathcal{L}_{\text{Image}}$ is defined as:
\begin{equation}
    \mathcal{L}_{\text{Image}} = \|\, \mathbf{M}_{\text{render}} - \mathbf{M}_{\text{2D}}\,\|_1.
\end{equation}

Second, the rasterized materials are used for deferred shading to generate a physically based rendering (PBR) image. This rendered image is then compared to the ground-truth input using the loss introduced in Gaussian Splatting~\cite{kerbl:23}. The rendering supervision loss is defined as:
\begin{equation}
    \mathcal{L}_{\text{3DGS}} = \lambda\, L_1(\mathbf{I}_{\text{PBR}}, \mathbf{I}_{\text{GT}}) + (1-\lambda)\, L_{\text{SSIM}}(\mathbf{I}_{\text{PBR}}, \mathbf{I}_{\text{GT}}),
\end{equation}
where $\mathbf{I}_{\text{PBR}}$ denotes the rendered image, $\mathbf{I}_{\text{GT}}$ is the ground-truth image, and $\lambda \in [0,1]$ (typically set to $0.8$). This loss supervises both the Neural Merger and the environment map estimation, ensuring that the final rendering is consistent with the input views.

\section{Experiments}

\begin{figure*}[t]
    \centering
    \includegraphics[width=\textwidth]{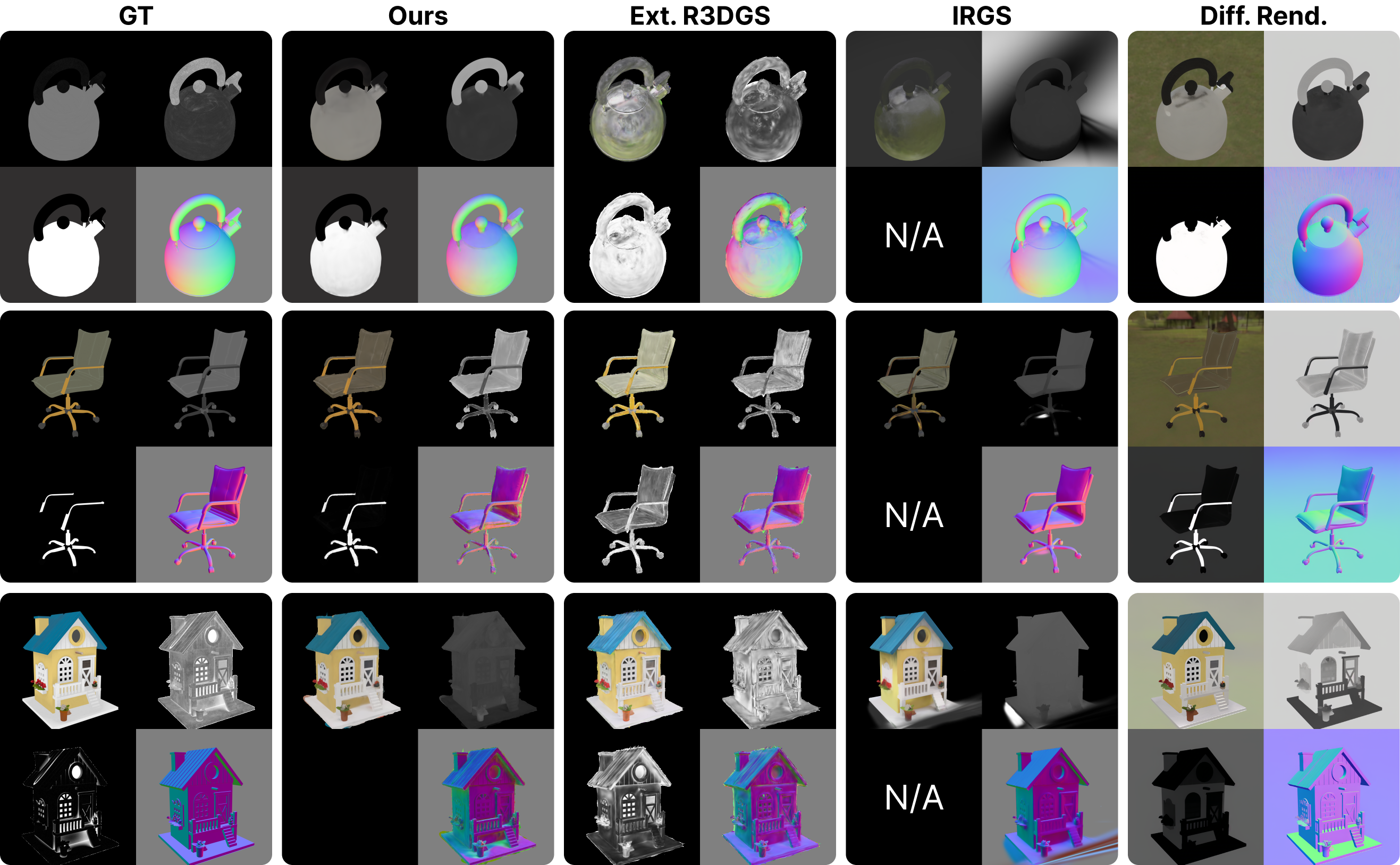}
    \caption{\textbf{Material Maps} produced by our method compared to extended R3DGS \cite{gao:24}, which can also predict metallic material maps, IRGS \cite{gu:25}, and the DiffusionRenderer material output produced on the test images that are not used for training. We show four images each, where the top left is the base color, top right is the roughness, bottom left is metallic, and bottom right are the normals. }
    \label{fig:G-Buffers}
\end{figure*}

\begin{figure*}[t]
    \centering
    \includegraphics[width=\textwidth]{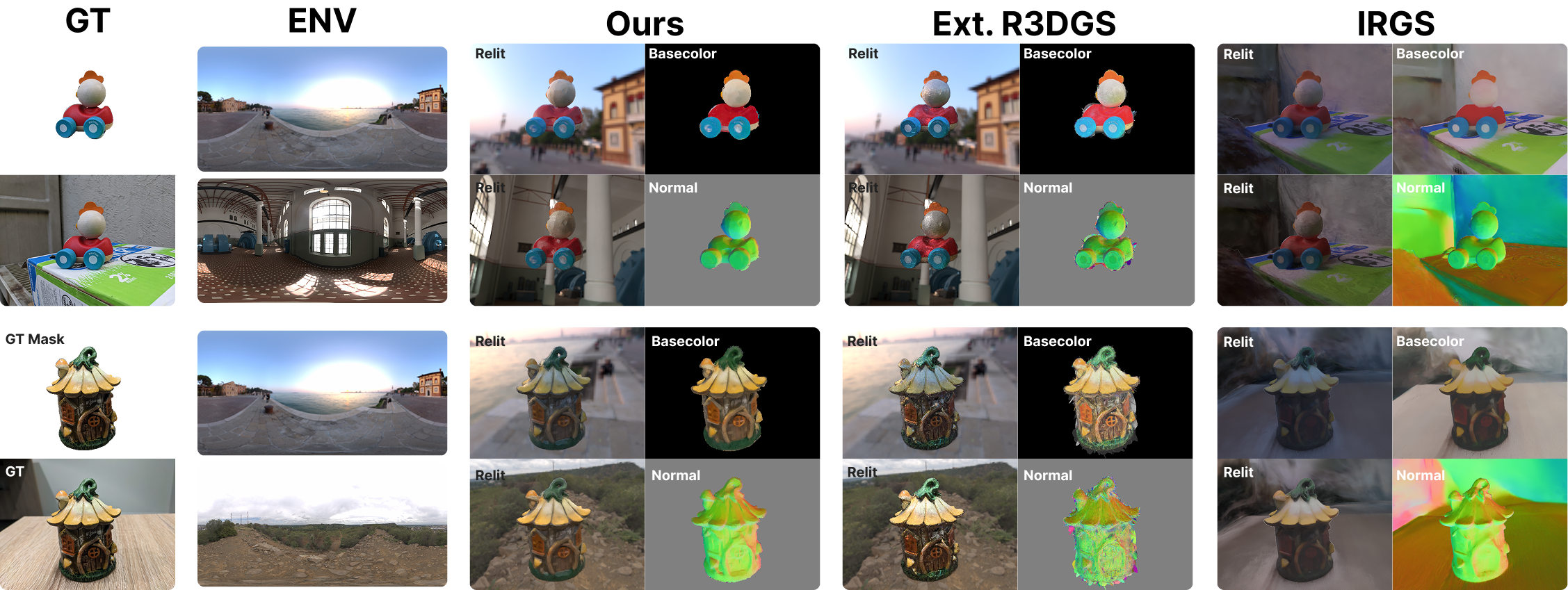}
    \caption{\textbf{Real-World Comparison} of our method, extended R3DGS \cite{gao:24} and IRGS \cite{gu:25}. The ground truth images show the object masked (top) and unmasked (bottom) to give a better understanding of the object and the surrounding lighting. }
    \label{fig:real_world}
\end{figure*}
We evaluate our method on both synthetic and real-world datasets from the Navi dataset~\cite{jampani2023navi}, comparing against state-of-the-art approaches for 3D material estimation and relighting, specifically an extended version of R3DGS~\cite{gao:24} (modified to support metallic materials) and IRGS~\cite{gu:25}. 
Our evaluation includes qualitative comparisons of material maps and relighting quality, quantitative metrics across material channels, computational performance, and an ablation study.

\subsection{Implementation Details}
\begin{table*}[t]
\centering
\caption{\textbf{Quantitative Comparison} of material estimation and relighting results on 17 synthetic objects. We compare against IRGS \cite{gu:25} and an extended version of R3DGS \cite{gao:24} that supports metallic materials. * For non-metallic objects, our model correctly optimizes the parameter to zero, which can result in infinite PSNR when all metallic maps are predicted as zero.}
\resizebox{\textwidth}{!}{
\begin{tabular}{l|ccc|ccc|ccc}
\toprule
Task & 
\multicolumn{3}{c|}{\textbf{Ours}} & 
\multicolumn{3}{c|}{\textbf{Ext. R3DGS}} & 
\multicolumn{3}{c}{\textbf{IRGS}} \\
\cmidrule(lr){2-10}
 & PSNR $\uparrow$ & SSIM $\uparrow$ & LPIPS $\downarrow$
 & PSNR $\uparrow$ & SSIM $\uparrow$ & LPIPS $\downarrow$
 & PSNR $\uparrow$ & SSIM $\uparrow$ & LPIPS $\downarrow$ \\
\midrule
Relighting & \cellcolor{bestcell}\textbf{27.282} & \cellcolor{bestcell}\textbf{0.897} & \cellcolor{bestcell}\textbf{0.080} & 25.483 & 0.875 & 0.094 & 24.409 & 0.850 & 0.166 \\
BaseColor  & \cellcolor{bestcell}\textbf{21.341} & \cellcolor{bestcell}\textbf{0.873} & \cellcolor{bestcell}\textbf{0.125} & 18.360 & 0.832 & 0.158 & 19.204 & 0.750 & 0.139 \\
Roughness  & 15.331 & \cellcolor{bestcell}\textbf{0.820} & \cellcolor{bestcell}\textbf{0.181} & 14.473 & 0.763 & 0.216 & \cellcolor{bestcell}\textbf{16.182} & 0.744 & 0.192 \\
Metallic   & \cellcolor{bestcell}\textbf{$\infty^*$/27.202} & \cellcolor{bestcell}\textbf{0.893} & \cellcolor{bestcell}\textbf{0.106} & 10.073 & 0.693 & 0.261 & N/A & N/A & N/A \\
\bottomrule
\end{tabular}
}
\label{tab:relighting_comparison}
\end{table*}

All methods use the same experimental setup for fair comparison. Synthetic scenes are initialized with a unit cube point cloud containing $100{,}000$ points sampled uniformly at random. Real-world scenes use structure-from-motion reconstruction from COLMAP~\cite{schonberger:16} for both initialization and camera inference. 

We evaluate on 17 synthetic objects with ground-truth material maps for quantitative analysis. The synthetic dataset uses $100$ training images and $200$ evaluation images per object. Real-world objects use all available images (average of 27 per object).

To handle highly specular objects, we train Gaussian Splatting on DiffusionRenderer normals as RGB, which helps guide the geometry more effectively, since Gaussian Splatting alone often struggles with specular surfaces and can produce holes in the reconstruction.

The Neural Merger consists of separate MLPs for each material channel (basecolor, roughness, metallic), each with 3 hidden layers of 128 neurons and ReLU activations. The final layer outputs view-specific weights passed through softmax to form a probability distribution. 

We spend 30.000 iterations for the 3D Gaussian geometry optimization and 10.000 for the material refinement.
All experiments run on an NVIDIA RTX 4090 GPU~\cite{nvidia_rtx_4090_specs_2022}.

We evaluate material estimation using PSNR, SSIM~\cite{wang2004ssim}, and LPIPS~\cite{zhang2018lpips} between predicted and ground-truth material maps. For relighting, we render novel views under different environment maps and compare against ground-truth renderings using the same metrics.

\subsection{Qualitative Results}

\paragraph{Relighting Quality}
Figure~\ref{fig:relighting} compares relighting quality across methods. Our method produces results that more closely resemble ground truth, particularly for specular objects. The extended R3DGS struggles with specular materials, often producing brighter images than ground truth due to unconstrained material maps during joint environment map optimization. In contrast, our method constrains material maps, enabling more accurate environment map optimization and improved material estimation.

IRGS exhibits artifacts such as floaters and overly flat surfaces. While it reconstructs flat surfaces well using 2D Gaussians, it loses fine detail compared to our method. Moreover, IRGS cannot predict metallic materials, preventing accurate reconstruction of highly specular metallic surfaces. Our method achieves clear advantages in reconstructing metallic and highly specular materials while preserving geometric detail across both diffuse and flat surfaces.

On the real-world relighting results in Figure~\ref{fig:real_world} one can observe the most consistent results with our method. With R3DGS, the estimated color appears too bright, the objects are too shiny and the geometry is less smooth, while IRGS produces smooth geometry but too dark colors. 

\paragraph{Material Maps} 
Figure~\ref{fig:G-Buffers} compares material maps across methods. Our method effectively removes baked-in lighting effects from basecolor maps, producing nearly diffuse basecolors with minimal shadowing, while other methods exhibit visible shadows and residual lighting.

Compared to DiffusionRenderer's original predictions, which exhibit significant view-dependent inconsistencies, our Neural Merger enhances multi-view consistency. This is particularly evident in roughness and metallic maps, where DiffusionRenderer's per-view predictions vary across views. Our approach removes these inconsistencies while preserving high-quality spatially varying material properties.

Our metallic maps match ground truth well, with predictions substantially closer than competing methods. For roughness, our method guides material maps toward consistent values for surfaces sharing the same properties, improving upon the 2D predictions. While DiffusionRenderer struggles with roughness accuracy (reflected in darker roughness maps), our Neural Merger refines initial predictions by enforcing view consistency, resulting in more physically plausible parameters. In contrast, R3DGS tends to overestimate roughness due to specular highlights appearing in only a subset of the images, biasing optimization toward diffuse surfaces. IRGS produces overly uniform roughness where fine details are difficult to discern.

Although IRGS normals are slightly closer to ground truth in some regions, our method significantly improves normals compared to extended R3DGS despite starting from the same 3D Gaussian geometry. Overall, our method produces qualitatively superior material maps across basecolor, normals, metallic, and roughness channels.

\subsection{Quantitative Results}

Table~\ref{tab:relighting_comparison} shows quantitative results on 17 synthetic objects. Our method consistently outperforms all baselines in relighting quality and basecolor estimation, consistent with the qualitative comparisons.

For roughness, IRGS achieves slightly higher PSNR (16.182 vs. 15.331), but our method achieves better SSIM (0.820 vs. 0.744) and LPIPS (0.181 vs. 0.192), indicating better preservation of structural information and perceptual quality. All methods face challenges in roughness estimation, which remains a difficult problem.

For metallic maps, our method shows substantial improvement. When correctly predicting fully non-metallic objects, PSNR becomes infinite, which occurs exclusively for our method. Those objects are left out in the actual PSNR calculation. While DiffusionRenderer often predicts partial metallicity in certain views our method enforces view-consistent material estimation, producing stable non-metallic predictions.

\subsection{Computation Time}

\begin{table}[b]
\centering
\caption{\textbf{Runtime Comparison} between our method and IRGS on the Navi dataset. All timings are reported in seconds and measured on an NVIDIA RTX 4090 GPU. Our method is approximately 3.5$\times$ faster than IRGS.}
\begin{tabular}{lcc}
\toprule
\textbf{Stage} & \textbf{Ours} (s) & \textbf{IRGS} (s) \\
\midrule
Diffusion Predictions & 112 & - \\
Gaussian Splatting & \cellcolor{bestcell}\textbf{131} & 2490 \\
Normal Generation (R3DGS) & 270 & - \\
Material Optimization & \cellcolor{bestcell}\textbf{975} & 2857 \\
\midrule
\textbf{Total} & \cellcolor{bestcell}\textbf{1488} & 5347 \\
\bottomrule
\end{tabular}
\label{tab:timings}
\end{table}

Table~\ref{tab:timings} shows runtime breakdown on the Navi dataset~\cite{jampani2023navi}. DiffusionRenderer requires $112$ seconds ($\sim$6 seconds per image) to process the full image set. Gaussian Splatting takes $131$ seconds on average (ranging from $64$ to $274$ seconds depending on object complexity). Normal generation using R3DGS takes $270$ seconds on average ($247$--$347$ seconds). Material optimization requires $975$ seconds on average, extending up to $3{,}631$ seconds (approximately one hour) for complex objects.

In total, our method takes $1{,}488$ seconds ($\sim$25 minutes) on average, approximately 3.5$\times$ faster than IRGS ($5{,}347$ seconds, $\sim$89 minutes).

\subsection{Ablation Study}
\begin{figure}[b]
    \centering
    \includegraphics[width=\columnwidth]{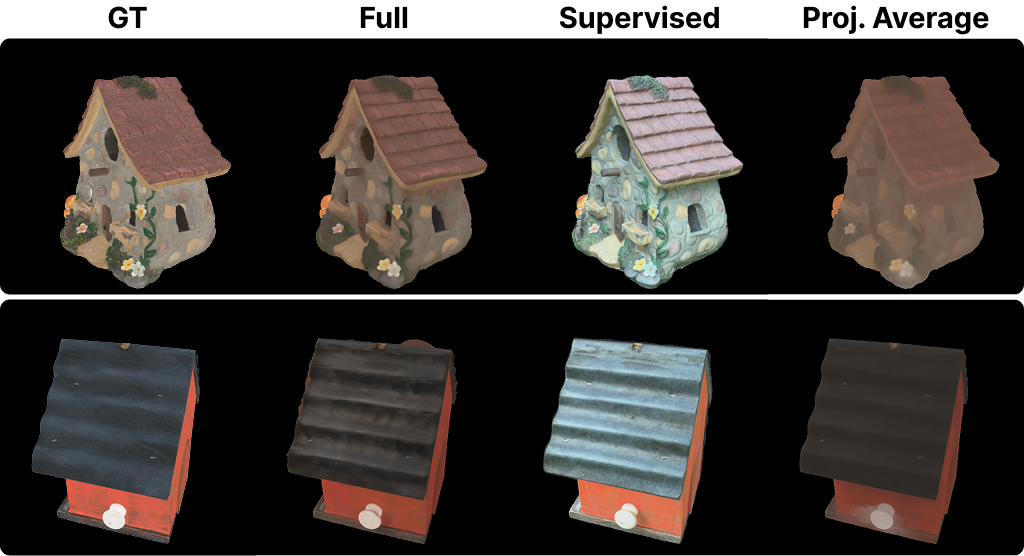}
    \caption{\textbf{Qualitative Comparison} of our ablations. The differences are most apparent in the base color, which has the largest impact on the qualitative appearance during relighting and consists of three channels rather than one.}
    \label{fig:ablation}
\end{figure}

\begin{table}[b]
\centering
\caption{\textbf{Quantitative Ablation} study on a subset of the evaluated objects. The \emph{Supervised} variant uses only the predicted 2D diffusion-based material maps for supervision, without any 3D optimization. The \emph{Proj. Average} variant directly projects all 2D predictions onto the Gaussians and averages them, without applying any further training or optimization. The \emph{Full} model includes the Neural Merger.}
\begin{tabular}{lccc}
\toprule
\textbf{Method} & \textbf{PSNR} $\uparrow$ & \textbf{SSIM} $\uparrow$ & \textbf{LPIPS} $\downarrow$ \\
\midrule
Full & \cellcolor{bestcell}\textbf{29.164} & \cellcolor{bestcell}\textbf{0.9105} & \cellcolor{bestcell}\textbf{0.0626} \\
Supervised          & 24.809 & 0.889 & 0.0792 \\
Proj. Average           & 25.555 & 0.866 & 0.122 \\
\bottomrule
\end{tabular}
\label{tab:ablation}
\end{table}

Table~\ref{tab:ablation} demonstrates that the Neural Merger is the key component responsible for our method's superior performance. The full model achieves the highest scores across all metrics (PSNR: 29.164, SSIM: 0.9105, LPIPS: 0.0626), significantly outperforming all ablated variants. This improvement stems from the Neural Merger's ability to enforce multi-view consistency while preserving high-quality material properties predicted by the diffusion model.

The \emph{Supervised} variant performs worse than the full model, relying solely on diffusion predictions without geometric or photometric optimization. Interestingly, it is even worse than the \emph{Proj.\ Average} baseline, which simply projects diffusion predictions into Gaussians without training. We attribute this to view-dependent effects captured in optimized Gaussians that are absent without optimization.

The qualitative results in Figure~\ref{fig:ablation} show that the Neural Merger produces cleaner and more consistent material maps. Visualized using base color (where baked-in lighting effects are most evident), the Neural Merger yields substantial improvements in both sharpness and color fidelity compared to the \emph{Proj.\ Average} baseline, which serves as our initialization. These results validate that the Neural Merger enhances visual quality and enforces consistency with underlying physical properties, resulting in more accurate and realistic relighting outcomes.

\section{Conclusion}
\label{sec:conclusion}

MatSpray enables casual acquisition and high-quality reconstruction of photorealistic relightable 3D assets with spatially varying materials.  
It effectively lifts 2D material predictions to 3D to fuse them with the 3D Gaussian geometry.
It employs pretrained 2D diffusion-based material estimators without requiring additional expensive training on large-scale 3D PBR data sets. 
Introducing the Neural Merger, our method significantly improves multi-view consistencies, which even video-based material prediction models still struggle with. 
The resulting relightable 3D models feature improved quality both in the estimated material maps as well as in the final relit appearance, even more so for highly specular objects. 
As demonstrated, MatSpray outperforms current state-of-the-art methods and excels in reconstructing accurate metallic maps for both synthetic and real-world inputs. The approach provides a powerful tool for easy 3D content generation.

\paragraph{Limitations}
While our approach drastically improves multi-view consistency, the overall material quality remains dependent on the performance of the chosen diffusion model. However, our PBR-to-image loss partially corrects small deviations in the diffusion predictions \ref{fig:G-Buffers} DiffusionRenderer vs. MatSpray. 

Our method struggles when inconsistent geometry and normals are produced by the underlying R3DGS method~\cite{gao:24}, though the photometric loss may partially mitigate these issues. Additionally, very small or flat Gaussians might sometimes be missed during ray tracing. Future work could address missing Gaussians through a projection transformer combination assignment similar to~\cite{ren2025mv}.

\paragraph{Future Work}
The high quality of the resulting 3D geometry-material association could be exploited for accurate 3D object part segmentation. This segmentation, paired with matching language features, might enable object-specific constraints in reconstruction or a natural interface for manipulating both geometry and reflections.

\begin{section}{Acknowledgements}
    \label{sec:acknowledgements}

    Funded by the Deutsche Forschungsgemeinschaft (DFG, German Research Foundation) under Germany's Excellence Strategy – EXC number 2064/1 – Project number 390727645.
    This work was supported by the German Research Foundation (DFG): SFB 1233, Robust Vision: Inference Principles and Neural Mechanisms, TP 02, project number: 276693517.
    This work was supported by the Tübingen AI Center.
    The authors thank the International Max Planck Research School for Intelligent Systems (IMPRS-IS) for supporting Jan-Niklas Dihlmann.
\end{section}

{
    \small
    \bibliographystyle{ieeenat_fullname}
    \bibliography{main}
}

\clearpage
\maketitlesupplementary
\section*{Overview}

This supplementary material provides extended results, analyses, and implementation details that complement the findings in the main paper.  
For ease of navigation, the main components are summarized here and referenced through the corresponding section labels.

\begin{itemize}
    \item \textbf{Additional Videos and Real-World Objects} (\ref{sec:videos}): A detailed collection of video comparisons and reconstructions of real objects that highlight the performance and stability of our method relative to earlier approaches.
    \item \textbf{Neural Merger Ablation} (\ref{sec:ablation}): An extended analysis of the importance of the final \texttt{Softmax} layer in the Neural Merger, supported by qualitative and quantitative evidence.
    \item \textbf{Tone Mapping Analysis} (\ref{sec:tone}): A discussion of the tone mapping behaviour of DiffusionRenderer, how this affects predicted base color, roughness and metallic maps, and why this creates a mismatch when compared to linear ground truth.
    \item \textbf{Implementation Details} (\ref{sec:implementation}): A description of our training setup, super sampling strategy, Neural Merger inputs and other practical considerations that are important for stable optimization.
\end{itemize}

\renewcommand\thesection{\Alph{section}}
\setcounter{section}{0}

\section{Additional Videos and Real-World Objects}
\label{sec:videos}
\begin{figure}[b]
\centering
\includegraphics[width=0.5\textwidth]{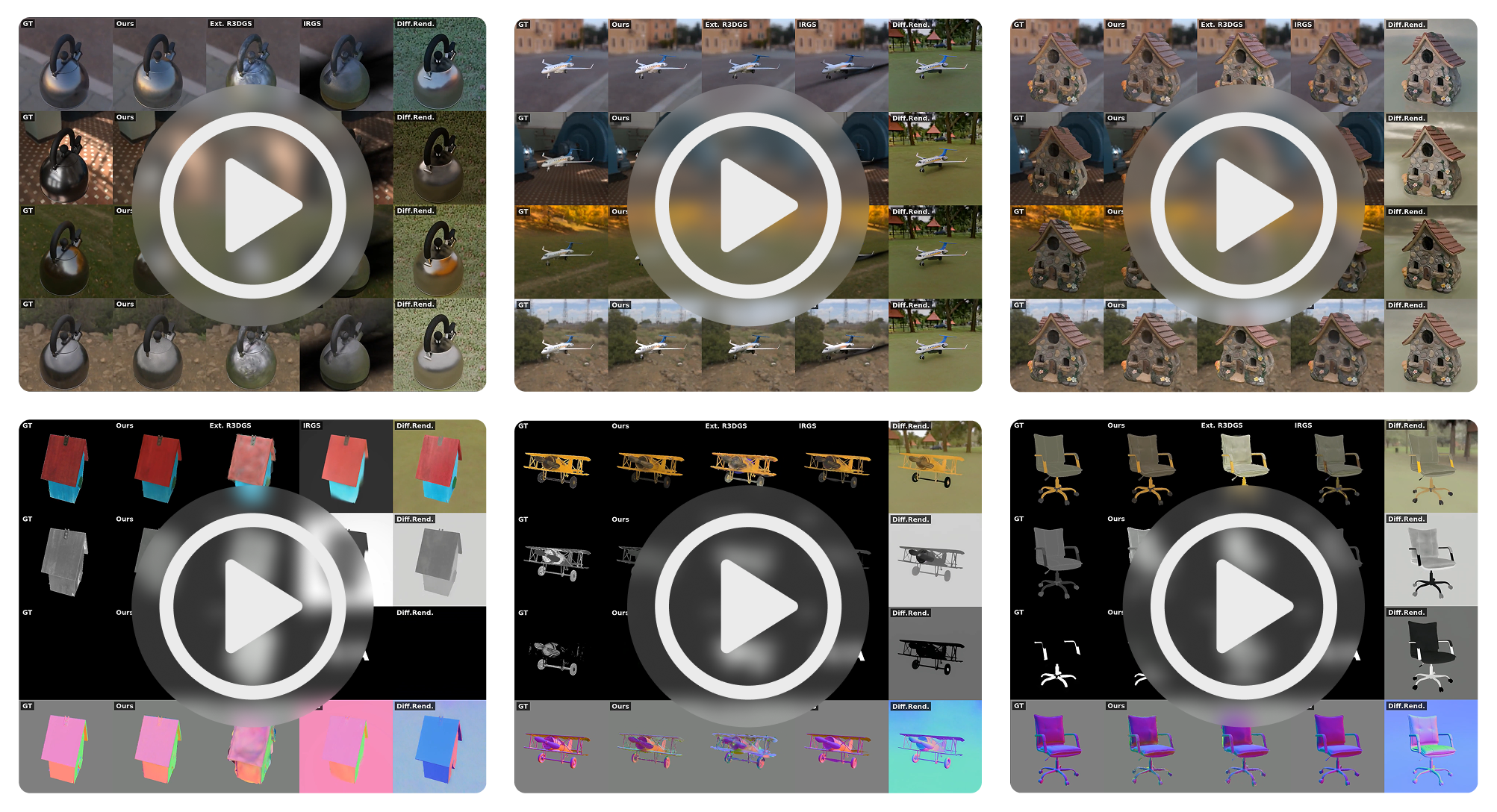}
\caption{Thumbnail showing six videos that can be viewed \href{https://drive.google.com/drive/folders/19RZpfUTo_Opdffv2gFX7mlsbniaVOp5w?usp=sharing}{here}. Three videos show relighting comparison and three show material prediction comparisons.}
\label{fig:thumbnail}
\end{figure}

\paragraph{}
Figure~\ref{fig:thumbnail} shows the thumbnail that links to all additional videos included with this supplementary material. These videos provide an extensive visual comparison of our method with Extended R3DGS \cite{gao:24}, IRGS \cite{gu:25}, and the forward renderer of DiffusionRenderer \cite{liang:25}. While the main paper presents representative examples, the extended videos give a more complete picture of the consistency and stability of our approach, especially compared to the produced material maps of DiffusionRenderer.

\paragraph{}
Across the set of videos, our method consistently produces reconstructions that remain stable across all viewpoints, without the flickering or structural collapse that can be observed in the other methods. This is particularly visible in objects with complex geometry or pronounced specular highlights such as the Kettle. Extended R3DGS often fails to maintain surface smoothness and yields unstable representations. On the other hand IRGS tends to oversmooth surfaces and tends to bake in specular reflections of metallic objects into its base color. In contrast, our approach maintains coherent structure even under strong lighting variations.

\paragraph{}
To illustrate this, we provide three relighting videos:  
\href{./Videos/airplane_whitegold_comparison.mp4}{White Golden Airplane},  
\href{./Videos/birdhouse_StoneHouse_comparison.mp4}{Stone Birdhouse}, and  
\href{./Videos/kettle_comparison.mp4}{Kettle}.  
Additionally, three videos visualize predicted material properties:  
\href{./Videos/airplane_yellow_materials_comparison.mp4}{Yellow Airplane},  
\href{./Videos/birdhouse_YellowFlower_materials_comparison.mp4}{Birdhouse with Yellow Flower}, and  
\href{./Videos/chair_materials_comparison.mp4}{Chair}.  
These examples show that DiffusionRenderer, despite being trained on its own dataset, still produces inconsistent material maps that vary strongly with camera angle and lighting. Our method mitigates these issues and aligns predictions across views more reliably.

\begin{figure*}[t]
\centering
\includegraphics[width=\textwidth]{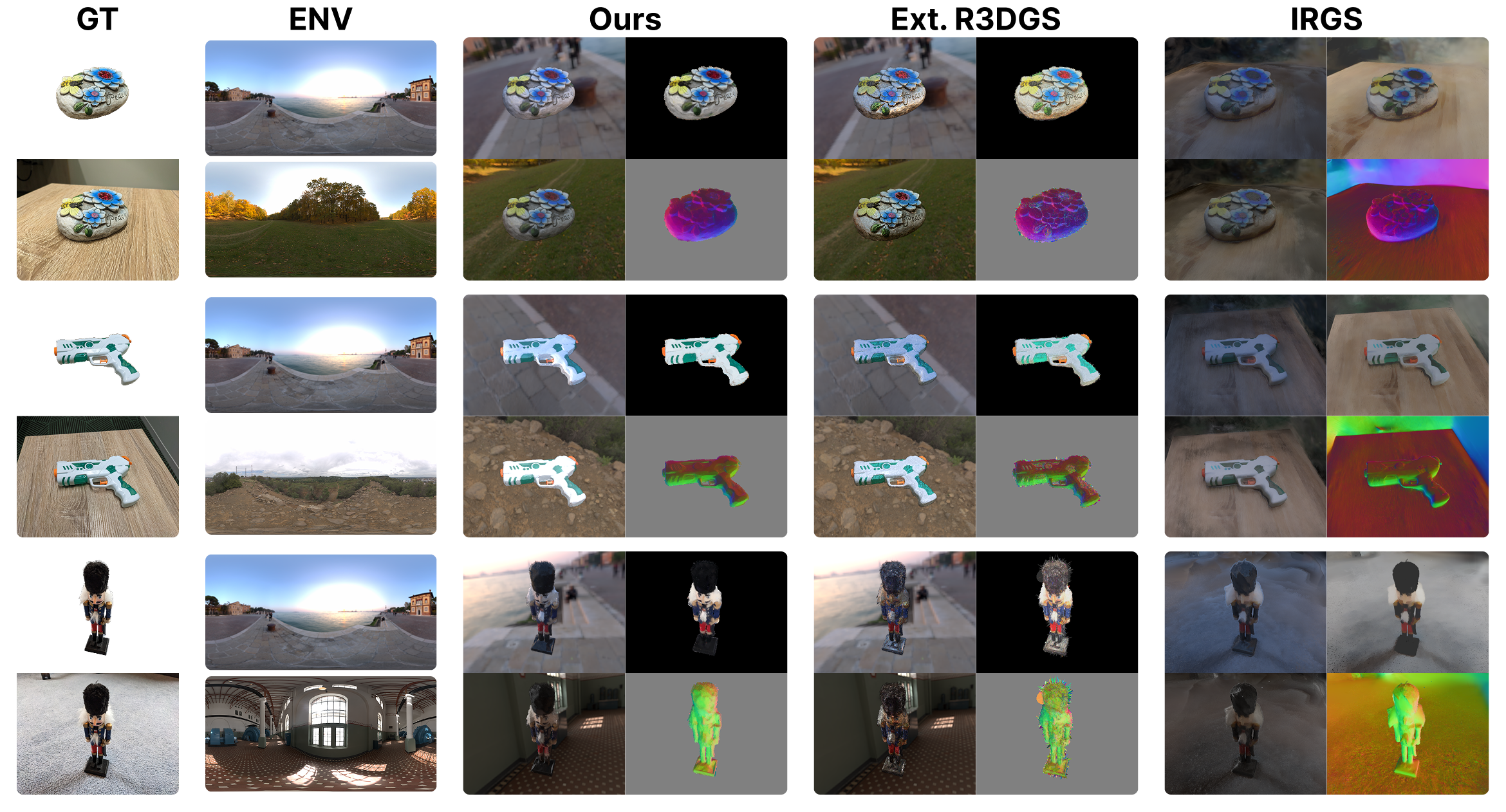}
\caption{Additional real objects reconstructed with our method, Extended R3DGS \cite{gao:24} and IRGS \cite{gu:25}. The figure includes relighting under two environments, base color and normal maps.}
\label{fig:additional_real_world}
\end{figure*}

\paragraph{Real-World Objects}
Figure~\ref{fig:additional_real_world} shows additional real objects reconstructed by our method and by Extended R3DGS and IRGS. Here, each method is evaluated under two relighting settings and compared in base color and normals. The differences are most obvious in the base color: our base color is locally sharp and coherent across the surface, while both baselines exhibit noise, distortions, or view-dependent artifacts. The relighting results further demonstrate that our predicted materials generalize well across lighting conditions, while the other methods still have lighting effects baked into their materials (R3DGS) or tend to be washed out (IRGS).

\begin{figure*}
\centering
\includegraphics[width=\textwidth]{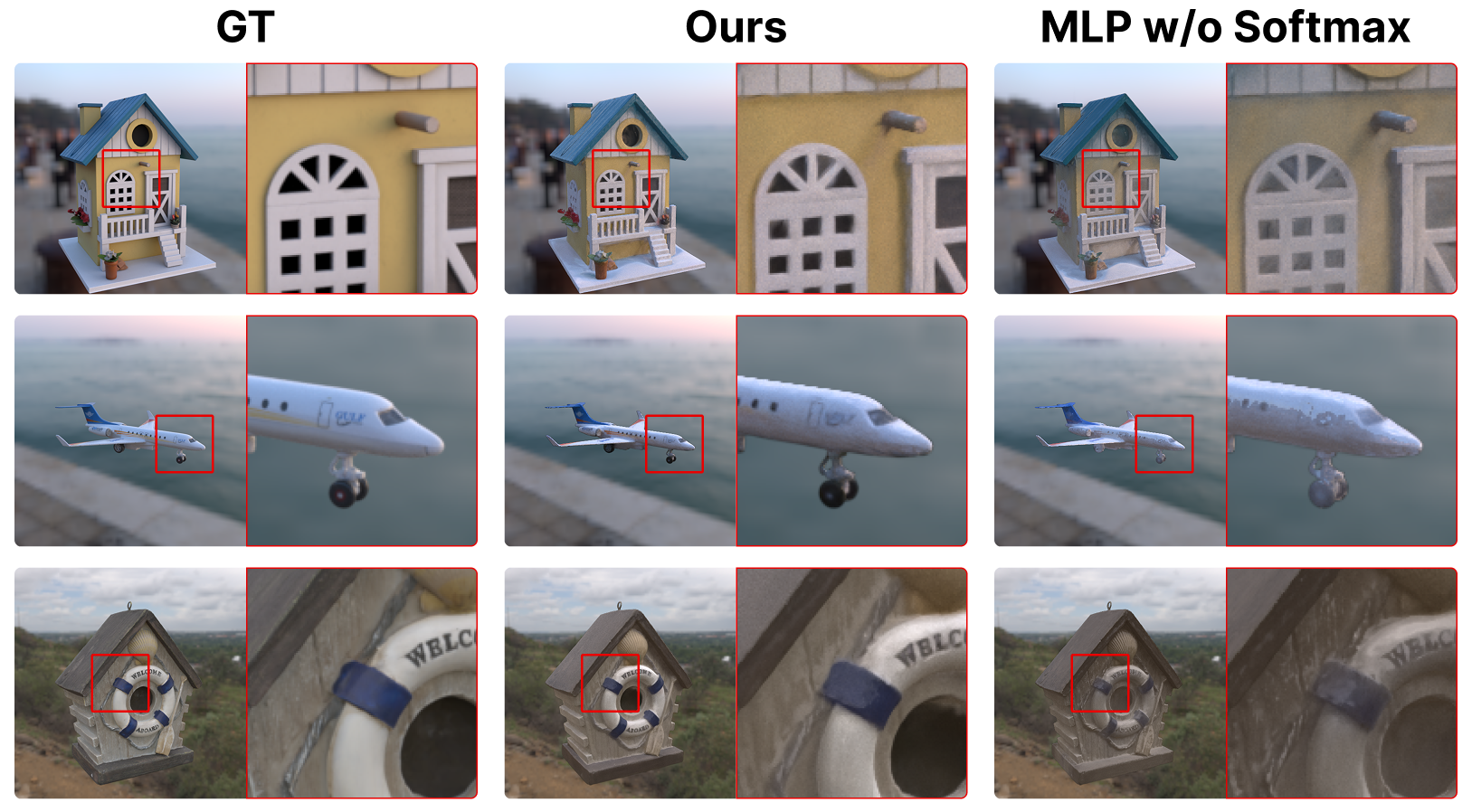}
\caption{The impact of the \texttt{Softmax} layer in the Neural Merger. Without it, lighting and shadow patterns leak into the material maps, leading to inconsistent relighting.}
\label{fig:mlp_softmax_ablation}
\end{figure*}

\begin{figure*}[t]
\centering
\includegraphics[width=\textwidth]{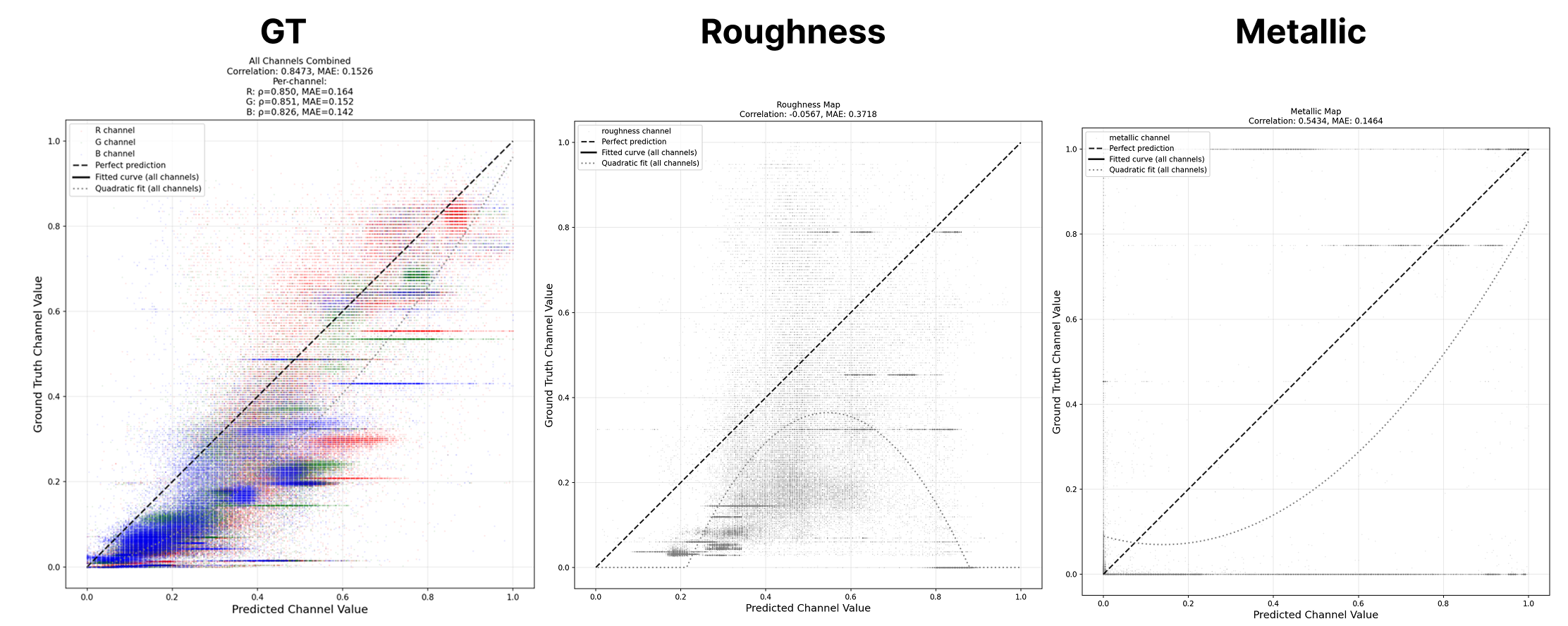}
\caption{Tone mapping applied by DiffusionRenderer significantly alters the appearance of material maps. The alpha mask removes background content and focuses on the region of interest.}
\label{fig:diffusionRenderer_analysis}
\end{figure*}
\section{Neural Merger Ablation}
\label{sec:ablation}

\paragraph{}
The Neural Merger plays a key role in ensuring that the material parameters assigned to each Gaussian remain stable and consistent across all viewpoints. One central element of the Neural Merger is the final \texttt{Softmax} layer, which normalizes its output into weights acting as a weighted average of the inputs. Although this layer may appear to be a small architectural detail, it has a sizable impact on the quality of the final reconstruction.

\paragraph{}
Without the \texttt{Softmax} normalization, the Neural Merger becomes unconstrained and starts to absorb illumination cues directly from the training images. In other words, instead of learning clean, view-independent materials, the MLP blends in signals that correspond to lighting variations and shadows. Because these patterns differ between viewpoints, the network produces material values that vary from view to view, which leads to inconsistency during rendering. Although this might also be additionally influenced by slight variations in the 2D Diffusion predictions geometry. This behaviour becomes especially problematic under relighting, because the embedded shadows and highlights interfere with the simulated lighting and produce unrealistic results.

\paragraph{}
Figure~\ref{fig:mlp_softmax_ablation} shows a comparison between the full method, the version without the \texttt{Softmax}, and the linear ground truth. The differences become clear when observing fine geometric structures and shadow placement. Without \texttt{Softmax}, shadows from the input images appear in the base color maps and the renderings become blurry in high detail areas. These issues are especially visible in the lower birdhouse example, where the version without \texttt{Softmax} fails to maintain consistent materials on the swim ring and the surrounding areas.

\paragraph{}
We further quantify these findings in Table~\ref{tab:ablation_softmax}, which reports results across all scenes in the dataset. The full model outperforms the version without the \texttt{Softmax} across all metrics, with especially large gains in perceptual similarity (LPIPS). This confirms that the Softmax-based normalization is not merely a numerical improvement but a key component that ensures robustness and prevents the network from encoding view-dependent appearance into the materials.

\begin{table}[b]
\centering
\caption{\textbf{Ablation Study} on all objects. Removing the Softmax layer causes the network to encode lighting, which degrades all metrics.}
\begin{tabular}{lccc}
\toprule
\textbf{Method} & \textbf{PSNR} $\uparrow$ & \textbf{SSIM} $\uparrow$ & \textbf{LPIPS} $\downarrow$ \\
\midrule
Full & \cellcolor{bestcell}\textbf{27.282} & \cellcolor{bestcell}\textbf{0.897} & \cellcolor{bestcell}\textbf{0.080} \\
Without Softmax & 24.600 & 0.874 & 0.114 \\
\bottomrule
\end{tabular}
\label{tab:ablation_softmax}
\end{table}

\section{DiffusionRenderer Tone Mapping Analysis}
\label{sec:tone}

\paragraph{}
One recurring observation in our experiments was that the base color predicted by our method tended to appear darker than the linear ground truth material map. This appeared to be a miss-prediction of the 2D material maps by DiffusionRenderer for a few objects. However, this discoloration appeared in almost all objects that we tested, hinting towards a systemic problem in DiffusionRenderer. Figure~\ref{fig:diffusionRenderer_analysis} illustrates this systemic discoloration of the predicted base color. This indicates that during training DiffusionRenderer was supervised using tone-mapped ground truth images.

\begin{figure}[b]
\centering
\includegraphics[width=0.5\textwidth]{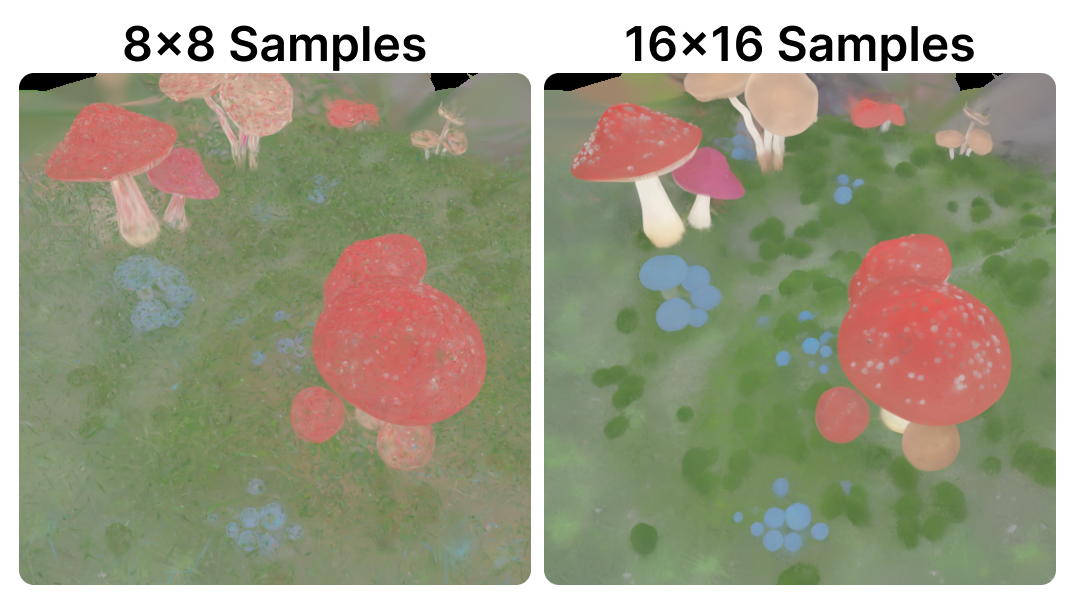}
\caption{Super sampling avoids missed Gaussians and ensures reliable projection of material supervision. Lower sampling rates cause holes and unstable geometry.}
\label{fig:super_sampling}
\end{figure}

\paragraph{}
Our analysis suggests that DiffusionRenderer employs a filmic or AgX tone mapping curve. These tone-mapping algorithms compress high dynamic range values into the range expected by standard displays, which improves visual quality but complicates the recovery of physically meaningful material parameters. In particular, these tone mappings are not analytically invertible, and even approximate inverse curves introduce errors, especially near shadows or highlights.

\paragraph{}
Base color is affected in a predictable way, because tone mapping acts like a softened gamma curve. Applying an inverse gamma of roughly one point eight partially recovers the linear values but cannot undo the full nonlinearity. Roughness is affected more severely, because its values occupy a small part of the zero to one interval, which collapses under tone mapping. Metallic values, on the other hand, remain closer to either zero or one and thus suffer less from compression. These effects explain why our predicted material maps sometimes differ from the linear ground truth as they closely match DiffusionRenderer’s tone-mapped output.
\section{Implementation Details}
\label{sec:implementation}

\paragraph{}
Our experiments were performed on an NVIDIA RTX 4090 GPU with PyTorch, C++ and Optix. To keep the input consistent with the internal resolution of DiffusionRenderer, we render all training views at a resolution of 512$\times$512 pixel. This choice ensures that the reconstruction quality aligns with the scale at which DiffusionRenderer was originally trained. In scenes with strong specular highlights, we disable geometry learning entirely and keep the Gaussian positions fixed, because additional geometric optimization tends to destabilize the representation under these conditions.

\paragraph{}
The Neural Merger is optimized using a learning rate of zero point zero zero one. Material supervision uses an L1-loss with a weight of 1.0, as we found that this balance prevents the model from overfitting shadows while still enforcing high fidelity in the material maps. During training, we also apply random view sampling to avoid biasing the model toward any particular viewpoint.

\paragraph{Super Sampling}
A key technical detail is the use of super sampling during the projection of material values into the Gaussian representation. We employ a 16$\times$16 grid of rays per pixel to ensure that even small or distant Gaussians receive material parameters. With fewer samples, Gaussians are occasionally missed leading to a patchy geometry and a low resolution material parameter transferal. Figure~\ref{fig:super_sampling} shows an example where a lower sampling rate produces obvious reconstruction defects.

\begin{figure}[b]
\centering
\includegraphics[width=0.25\textwidth]{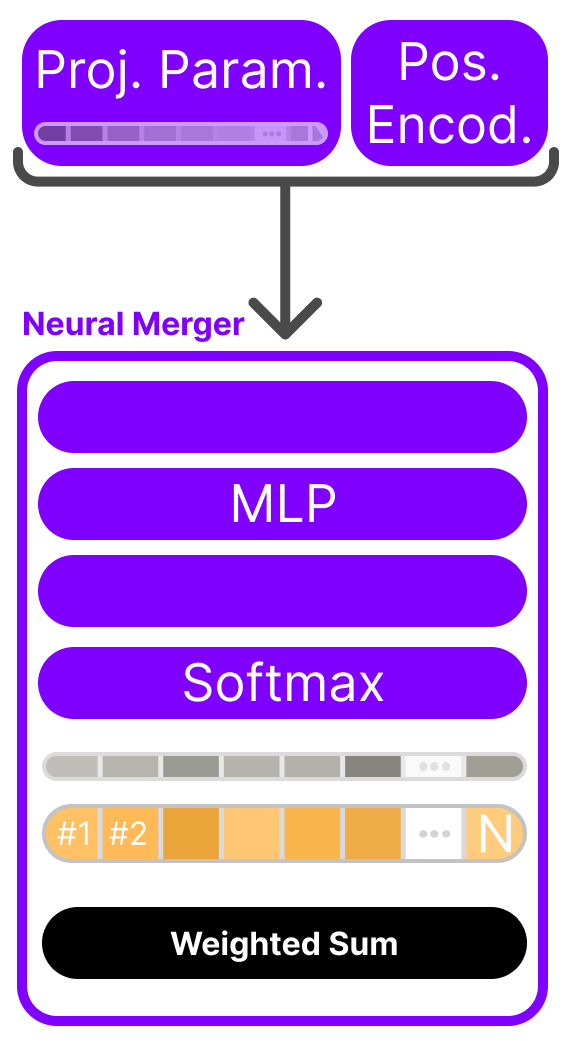}
\caption{The input to the Neural Merger includes positional codes and projected material parameters.}
\label{fig:mlp_input}
\end{figure}

\paragraph{Merger Inputs}
Finally, Figure~\ref{fig:mlp_input} illustrates the input to the Neural Merger. The features consist of a NeRF-style positional encoding of the Gaussian location along with the projected base color, roughness and metallic values. The combination of positional encoding and projected materials allows the network to balance local detail with global consistency, which is essential for producing clean results under relighting.

\end{document}